\pdfoutput=1
\documentclass[lettersize,journal]{IEEEtran}
\usepackage{amsmath,amsfonts}
\usepackage{float}
\usepackage[ruled]{algorithm2e}
\usepackage{array}
\usepackage[caption=false,font=scriptsize,labelfont=sf,textfont=sf]{subfig}
\usepackage{textcomp}
\usepackage{url}
\usepackage{verbatim}
\usepackage{graphicx}
\usepackage{cite}
\usepackage{xcolor}
\usepackage{setspace}
\usepackage{bm}
\usepackage{multirow}
\usepackage{soul}
\allowdisplaybreaks
\usepackage{booktabs}
\usepackage{pifont}
\usepackage{colortbl}
\definecolor{lightblue1}{RGB}{220,230,242}
\definecolor{lightblue2}{RGB}{235,245,255}
\usepackage{makecell}
\usepackage{tabularx}
\usepackage{dblfloatfix}
\usepackage{tcolorbox}
\tcbuselibrary{listingsutf8}
\usepackage{listings}
\title{Adopting Reinforcement Learning with Verifiable Rewards for Molecular Generation}

\author{Mingxuan Ouyang, Hao Lan, Wanyu Lin}

\begin{document}
\maketitle

\begin{abstract}Leveraging large language models (LLMs) for molecular generation has shown remarkable potential in chemical and drug design. Current methods primarily rely on supervised training or fine-tuning with limited datasets, which are insufficient to capture complex molecular design objectives. While some approaches attempt to guide generation toward specific goals, they often lack direct optimization mechanisms, making it difficult to align generated molecules with desired properties.
To tackle these challenges, we propose \textbf{LLMol}, a principled reinforcement learning framework that directly incorporates verifiable rewards for targeted molecule generation. The key insight is to formulate molecular design as a goal-conditioned sequence prediction task, where verifiable rewards serve as explicit supervision to drive generation toward desired objectives.
LLMol follows a two-stage training paradigm combining supervised learning and reinforcement learning. In the first stage, large language models are supervised fine-tuned to capture chemical syntax and molecular distributions. In the second stage, we introduce Reinforcement Learning with Verifiable Rewards (RLVR), which directly integrates property-based reward signals to guide molecular generation toward task-specific objectives.
To address the high variance and instability common in discrete sequence optimization, we adopt Group Relative Policy Optimization (GRPO), a stable on-policy algorithm that smooths reward signals and improves training robustness. This framework enables LLMol to effectively handle a range of molecular design tasks, including single-property targeting (e.g., penalized logP, QED) and structure-constrained optimization.
Experimental results demonstrate that LLMol consistently outperforms existing methods, achieving higher success rates and improved efficiency across diverse molecular benchmarks.
\end{abstract}

\section{Introduction}

\label{intro}
Molecular generation aims to automatically find novel chemical structures that satisfy a desired molecular profile, being increasingly important in various applications such as drug discovery and material~\cite{xiecrystal, xu2022geodiff, hongaccelerating, hong2024diffusion, wang2024crystalline}. This task is challenging due to the vast and discrete nature of molecular space, estimated to be as large as $10^{33}$, making exhaustive searches impractical~\cite{polishchuk2013estimation}. Exhaustive exploration of such a vast space is computationally impractical, motivating the development of generative models that can efficiently navigate and exploit the underlying chemical landscape. 
Traditional approaches often rely on high-throughput virtual screening~\cite{macarron2011impact}, discrete local search methods~\cite{rupakheti2015strategy} or similar interpolation techniques. 
These processes are not only time-consuming and labor-intensive but also require substantial financial resources.

In recent years, deep learning has emerged as a promising approach for molecular design~\cite{macedo2024medgan, xiecrystal, xu2022geodiff, song2024equivariant}. There has been an increasing focus on developing deep learning algorithms capable of automatically generating chemically valid molecular structures while simultaneously optimizing their properties. 
Several deep generative models have been explored to identify promising drug candidates~\cite{kingma2013auto, gomez2018automatic, jin2018junction, eckmann2022limo, guimaraes2017objective, kadurin2017drugan, you2018graph, luo2021graphdf, shi2020graphaf}. These models, often trained on extensive molecular datasets, are adept at learning general chemical distributions and generating valid molecular structures. Many approaches seek to optimize these models toward specific properties by incorporating additional property predictors or relying on expert-defined scoring functions~\cite{jin2018junction, gomez2018automatic, liu2018constrained}. However, aligning molecule generation with complex chemical objectives remains a significant challenge, especially when optimizing multiple properties or substructures simultaneously.  Moreover, these approaches typically operate in a fixed-dimensional latent space, which may limit their capacity to capture the full diversity of chemical structures~\cite{wang2022retrieval}.

Recent advancements in large language models (LLMs) have demonstrated remarkable capabilities in learning from vast corpora and generating coherent, contextually relevant sequences across a wide range of tasks~\cite{qwen}. By treating molecular strings as a "chemical language," LLMs have been successfully applied to achieve high-throughput and accurate molecular generation~\cite{flam2022language}. For example, string-based representations like SMILES~\cite{weininger1988smiles} have gained significant attention in molecular language models due to their simplicity and adaptability~\cite{deng2023systematic, wigh2022review}.
By learning from molecular string datasets, LLMs can capture complex chemical rules and syntactic patterns, enabling them to generate valid and diverse molecular structures~\cite{flam2022language, irwin2022chemformer}. Recent studies have demonstrated the potential of LLMs in string-based molecular generation~\cite{li2024empowering, liu2024conversational}.

Despite recent progress, existing LLM-based methods for molecular generation primarily rely on supervised fine-tuning (SFT) or in-context learning to mimic distributions observed in training data~\cite{wang2022retrieval}. These approaches often fall short in exploring the vast chemical space or in generating molecules with specific, desirable properties. To address this, reinforcement learning (RL) has been introduced to steer generation toward task-specific objectives~\cite{fang2024domain, popova2019molecularrnn, luo2021graphdf, shi2020graphaf}. While effective in principle, many RL-based methods either lack direct optimization mechanisms for aligning generation with desired molecular properties, or suffer from high variance in reward signals and unstable training dynamics—issues that are particularly pronounced in discrete sequence generation~\cite{xie2021mars}.

To address these challenges, we propose a novel molecular design framework based on Reinforcement Learning with Verifiable Rewards (RLVR)~\cite{lambert2024t}. 
Following the standard two-stage paradigm commonly adopted in LLM training~\cite{ouyang2022training}, we first apply supervised fine-tuning with LoRA~\cite{hu2021lora} on a pre-trained language model to capture chemical syntax and the underlying molecular distribution. 
Building on this foundation, we perform reinforcement learning using property-based rewards computed via RDKit~\cite{landrum2013rdkit}, ensuring that reward signals are objective, reproducible, and aligned with the target molecular properties.
Unlike supervised learning or preference optimization approaches that require labeled datasets, LLMol generates synthetic training data online by sampling from the evolving policy during optimization. 
To further address the high variance typically observed in reinforcement learning over discrete sequences, we adopt Group Relative Policy Optimization (GRPO)~\cite{shao2024deepseekmath}, a stable on-policy algorithm that smooths reward signals and mitigates the bias introduced by off-policy methods.
Our approach effectively bridges the gap between chemical language modeling and goal-directed molecular optimization, providing a robust and scalable solution for generating molecules with target-specific properties.

We demonstrate the effectiveness of our framework by evaluating it across various molecular design tasks, including: (i) generating molecules with specified target properties (e.g., logP) within desired numerical ranges; (ii) exploring molecular spaces to maximize key properties such as QED and penalized logP; and (iii) optimizing target properties under the constraint of a given reference structure. Experimental results demonstrate that our method outperforms state-of-the-art baselines across all the aforementioned molecular generation tasks. Our major contributions are summarized as follows:

\begin{enumerate}

    \item We propose a novel two-stage training framework that combines supervised fine-tuning and reinforcement learning with verifiable rewards. The model is first adapted to the molecular domain by learning general chemical syntax and distribution through SFT, and then further optimized for task-specific property objectives using deterministic reward signals. Leveraging LoRA, the framework enables efficient and modular adaptation across diverse molecular generation tasks. 

    \item To address the high variance in RL training, we introduce Group Relative Policy Optimization, an on-policy optimization method that smooths reward signals and enables more stable and efficient exploration of the chemical space. 

    \item We comprehensively validate our framework across diverse molecular design tasks, demonstrating its capability in producing chemically valid molecules, navigating chemical spaces efficiently, and achieving notable optimization.

\end{enumerate}

\section{Related work}

\label{related_work}
\subsection{Molecule generation models.} 
De novo molecular design aims to generate novel chemical structures with desired properties, rather than merely searching within existing molecular databases~\cite{gomez2018automatic, jin2018learning}. In recent years, deep generative models have emerged as powerful tools, offering optimization pipelines for exploring focused sets of molecules with improved performance~\cite{elton2019deep}. 
One popular direction involves learning continuous latent representations of molecules through models such as variational autoencoders~\cite{eckmann2022limo,baldi2012autoencoders, jin2018junction,winter2019efficient}. These methods encode discrete molecular inputs into continuous spaces where gradients can be used to optimize properties, often in combination with Bayesian optimization~\cite{gomez2018automatic,winter2019efficient}. However, these approaches are sensitive to the quality and geometry of the latent space, and mapping optimized latent vectors back to valid molecules remains a non-trivial challenge~\cite{jin2018junction}.
Generative adversarial networks have been shown to generate diverse libraries of realistic molecules with specific properties~\cite{goodfellow2020generative, putin2018reinforced, mendez2020novo}. Some graph-based approaches directly train a translation model to map input molecules to high-quality output molecules~\cite{jin2020hierarchical}. While conceptually straightforward, these methods depend heavily on large volumes of high-quality labeled data, which limits their applicability in data-scarce settings. The evolutionary algorithms and genetic algorithms~\cite{tripp2023genetic, nigam2019augmenting} apply biologically inspired search strategies to iteratively improve molecular candidates through mutation and crossover operations. While flexible and domain-agnostic, such methods often rely on handcrafted rules and require extensive expert knowledge to design effective mutation strategies.

Recent developments in advanced natural language processing, especially through the adoption of transformer-based architectures, have been leveraged in molecular design or molecular optimization~\cite{flam2022language, irwin2022chemformer, born2023regression, edwards2022translation, liu2024conversational, fang2024domain}. These models have proven effective in generating simplified molecular-input line-entry system (SMILES) representations~\cite{weininger1988smiles}, facilitating the systematic exploration of novel chemical compounds. By leveraging large-scale pretraining, multitask learning, and retrieval-augmented generation, these models can be adapted for conditional or property-guided molecule generation. Their compatibility with string-based representations and modular fine-tuning pipelines makes them highly scalable and easy to integrate into existing workflows.
However, most existing language model based approaches rely heavily on supervised learning from limited molecular datasets. As a result, they primarily model known distributions of existing molecules, which significantly constrains their ability to generate novel compounds with rare or optimized properties.

\subsection{Reinforcement learning-based molecule generation} 

Reinforcement learning has emerged as a critical paradigm for post-training generative models to better align with task-specific objectives~\cite{liang2021hierarchical, ouyang2022training}.
In natural language tasks, RL from human feedback (RLHF) has been widely used to improve model helpfulness and safety~\cite{o2024open, ouyang2022training}, but it often relies on learned reward models, which may be susceptible to overoptimization and misalignment~\cite{gao2023scaling}. 
In contrast, recent systems such as DeepSeek-R1~\cite{guo2025deepseek, shao2024deepseekmath} and Tülu 3~\cite{lambert2024t} adopt reinforcement learning with verifiable rewards (RLVR), leveraging rule-based or programmatic reward functions that provide deterministic and objective feedback. This framework has demonstrated strong performance in domains such as math, code, and reasoning. In molecular generation, RLVR is particularly suitable due to the availability of domain-specific toolkits (e.g., RDKit) that can evaluate molecular validity, property thresholds, and structural similarity. Compared to RLHF, RLVR eliminates the reliance on learned reward models, offering more consistent and objective feedback for molecular design.

However, RL-based molecular generation model~\cite{popova2018deep, popova2019molecularrnn, shi2020graphaf} still faces challenges such as high variance and unstable optimization, especially in discrete sequence settings~\cite{xie2021mars}. Proximal Policy Optimization (PPO)~\cite{schulman2017proximal} is a commonly used algorithm that mitigates variance via a learned value function (critic). While effective, PPO introduces substantial memory and computational burden, as the critic model is often comparable in size to the policy. Moreover, the rewards in LLM-based molecular generation tasks are typically sparse and delayed, as they are only available after the complete molecular sequence has been generated. For many chemical properties, (e.g., QED), it is extremely challenging to decompose the final reward into meaningful token-level contributions. As a result, learning a stable and accurate value function that operates at each decoding step becomes particularly difficult. 
Group Relative Policy Optimization (GRPO)~\cite{shao2024deepseekmath, mroueh2025reinforcement} addresses these limitations by removing the critic model and instead using group-based Monte Carlo rollouts to estimate relative advantages. Rewards are standardized through whitening, improving training stability and facilitating better exploration. GRPO complements RLVR by naturally supporting verifiable, deterministic rewards without requiring additional learned components.

\begin{table*}[!t]
\centering
\small
\caption{Comparison of average tokenization lengths for different molecular representations on ZINC250K.}
\label{tab:token-length-comparison}
\renewcommand{\arraystretch}{1.3}
\begin{tabularx}{\textwidth}{@{}l >{\raggedright\arraybackslash}X c@{}}
\toprule
\textbf{Representation} & \textbf{Example} & \textbf{Avg. Length} \\
\midrule
SMILES & [NH3+][C@@]1(O)CCc2ccccc21 & 30.46 \\
SELFIES & 
[NH3+1][C@@][Branch1][C][O][C][C][C][=C][C][=C][C][=C][Ring1][=Branch1][Ring1][\#Branch2]
& 99.36 \\
Simplified SELFIES & 
NH3+1 C@@ Branch1 C O C C C =C C =C C =C Ring1 =Branch1 Ring1 \#Branch2
& 61.93 \\
\bottomrule
\end{tabularx}
\end{table*}

\section{Preliminaries}

\subsection{Molecule representation} 
String-based representations are widely adopted for molecular generation tasks due to their compatibility with language models. Among these, SMILES (Simplified Molecular Input Line Entry System)~\cite{weininger1988smiles} and SELFIES (Self-Referencing Embedded Strings)~\cite{krenn2020self} are the most commonly used. 
SMILES possesses the advantage of conciseness but struggles with generating both syntactically and semantically valid strings. In contrast, SELFIES is a fully robust molecular language that ensures both syntactic and semantic validity by design.
However, the use of square brackets in SELFIES leads to longer token sequences when processed by language models, often increasing vocabulary complexity and introducing tokenization inefficiencies. To mitigate this, we adopt a simplified SELFIES representation that removes square brackets and inserts spaces between semantic units. This transformation brings the representation closer to natural language syntax and reduces token length, thus improving the efficiency of LLM-based molecular modeling. Table~\ref{tab:token-length-comparison} presents an example of each representation and compares their average tokenization lengths on the ZINC250K dataset~\cite{sterling2015zinc}.

\subsection{Task description}
\label{sec: task-description}
In molecular design for drug discovery, various generative tasks are employed to navigate the vast chemical space and construct candidate molecules that meet specific optimization goals, such as bioactivity, drug-likeness, or synthetic accessibility. In this work, we consider three representative molecule generation scenarios following LIMO~\cite{eckmann2022limo}:

\textbf{logP targeting:} This task involves generating molecules whose logP (octanol-water partition coefficient) falls within a specified numerical range. The logP value reflects a molecule's hydrophilicity or lipophilicity and is a critical determinant of absorption, distribution, and permeability—key aspects of pharmacokinetics. By constraining logP to a narrow interval, this task simulates early-stage filtering for drug-likeness based on physiochemical criteria, and evaluates a model's ability to precisely control continuous-valued molecular properties.

\textbf{QED and penalized logP maximization:} This task focuses on generating molecules with high scores in penalized logP (P-logP) or the quantitative estimate of drug-likeness (QED). The P-logP metric is derived from the estimated octanol-water partition coefficient (logP), penalized by the synthetic accessibility (SA) score and the number of cycles with more than six atoms~\cite{jin2018junction}. QED quantifies drug-likeness based on a combination of physicochemical properties~\cite{bickerton2012quantifying}. These properties are important considerations in drug discovery, and this task evaluates the model's ability to optimize salient molecular attributes, even if the maximization of these metrics in isolation may be of limited utility~\cite{zhou2019optimization}.

\textbf{Similarity-constrained target property maximization:} Given a set of starting molecules, this task focuses on generating novel molecules that achieve higher P-logP or QED scores while maintaining structural similarity to the original compounds. Similarity is quantified using the Tanimoto coefficient computed over Morgan fingerprints~\cite{rogers2010extended}. This setting reflects a common scenario in drug discovery, where the objective is to optimize molecular properties without straying too far from a known active compound, thus increasing the likelihood of preserving biological activity.

\subsection{Prompt design} 
To train models that can be used for above tasks, we use task-specific prompts. The input to the model is an instruction prompt followed by the simplified SELFIES. The prompt of each task comprises three components: a task description, illustrative examples, and a generation instruction. The inclusion of examples serves to enhance the model's stability on molecular tasks and to reduce hallucinations during generation. Moreover, to emphasize key information and help the model distinguish between the instruction and the target molecule, we enclose relevant strings with closed tags such as \texttt{<value></value>} and \texttt{<SELFIES></SELFIES>}. To improve robustness, we construct multiple semantically equivalent prompt templates per task. These variants are randomly sampled during training to prevent overfitting to prompt phrasing. Due to space constraints, the full prompt formatting table is provided in the appendix~\ref{appendix:prompt}.

\subsection{Problem definition}
Our objective is to fine-tune LLMs to generate molecular strings that align with desired properties and structural constraints. LLMs perform next-token prediction over sequences, modeled as a categorical distribution \( p(w_{t+1} | w_{0:t}) \), where \( w_0:t \) represents the input token sequence (the prompt), and \( w_{t+1} \) is the predicted next token. To generate a target molecule \( m \), the desired properties \( y_i \) and constraints \( s_i \) are encoded into a conditional instruction prompt \( c \). We aim to fine-tune a pre-trained large language model \( \text{LLM}_{\text{pre}} \) to capture the conditional distribution \( p(m|c) \), ensuring that the generated molecule satisfies both the desired properties \( y_i \) and constraints \( s_i \) described in the prompt. The generated molecule \( m \in \mathcal{S} \) is represented as a simplified SELFIES string over vocabulary \( \mathcal{V} \), and the model operates autoregressively, generating one token at a time until an end-of-sequence token is produced.

\section{Method}
\label{method}

\begin{figure*}[!t]
\centering
\includegraphics[width=\textwidth]{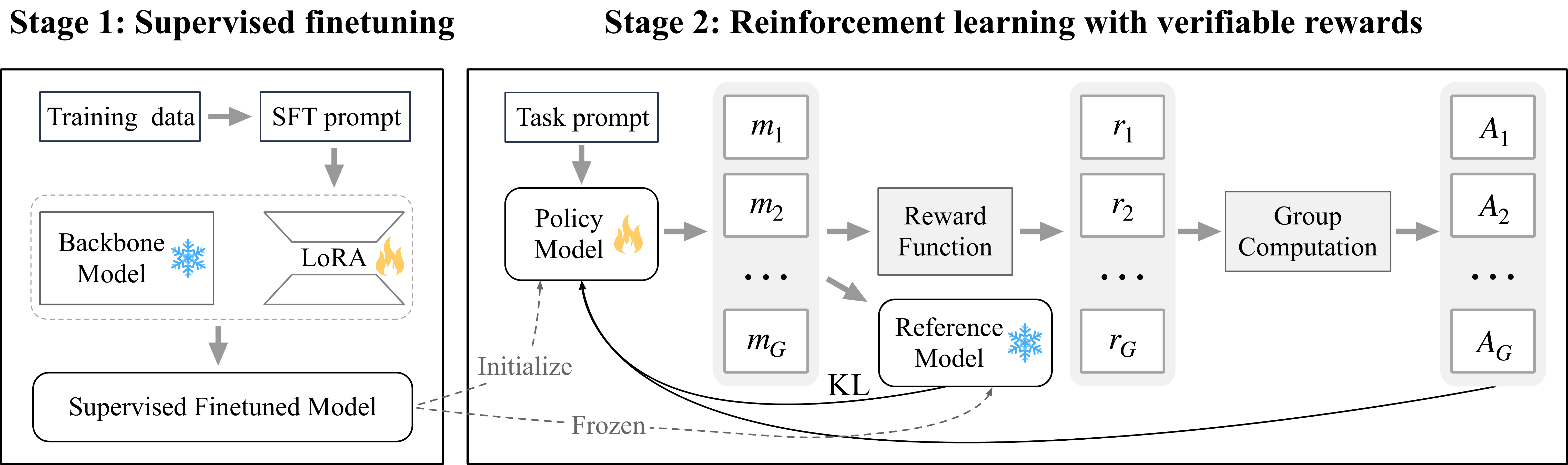}

\caption{
An illustration of the proposed LLMol framework for one molecular generation task. Stage 1 performs supervised fine-tuning on a pretrained backbone language model using LoRA to obtain a chemically-aware generator. The Policy Model in Stage 2 is initialized from the supervised finetuned model. Given a task prompt, the model generates candidate molecules, which are scored by a deterministic reward function and optimized via Group Relative Policy Optimization using group-level statistics. A KL divergence penalty with respect to the frozen reference model stabilizes training. The framework supports multi-objective optimization across diverse molecular design tasks via task-specific LoRA adapters and flexible prompt conditioning. Modules marked with a \textcolor[RGB]{255,204,102}{\textbf{Flame}} symbol contain trainable parameters, while those marked with a \textcolor[RGB]{117,196,254}{\textbf{Snowflake}} symbol are frozen.
}

\label{fig:framework}
\end{figure*}

We present LLMol, a two-stage molecular generation framework that includes supervised fine-tuning and reinforcement learning with verifiable rewards. As illustrated in Fig.~\ref{fig:framework}, Stage 1 performs supervised fine-tuning with LoRA to adapt a pretrained language model for valid molecular generation; Stage 2 applies reinforcement learning with verifiable rewards, where the policy is optimized using Group Relative Policy Optimization to satisfy property-driven design goals.
The LLMol framework supports multiple molecular design tasks, including logP targeting, QED and penalized logP maximization, and similarity-constrained optimization. Each task is defined by a distinct reward function and is instantiated via prompt conditioning and task-specific LoRA adapters. In the following sections, we detail each training stage and the reward design strategies in turn.

\subsection{Supervised finetuning}
We adopt a decoder-only causal language model as the base generator, leveraging its strong autoregressive modeling capabilities and architectural flexibility.  Molecular generation is framed as a conditional sequence modeling task, where the model generates a token sequence representing a molecule, conditioned on a prompt $c$ that encodes desired properties or structural constraints.

To adapt the base model to molecule generation, we perform supervised fine-tuning (SFT) using cross-entropy loss. As illustrated in Stage 1 of Fig.~\ref{fig:framework}, the input consists of a concatenated prompt–molecule pair, but the loss is only applied over the molecule tokens. Let $m^* = \{t_1^*, \ldots, t_l^*\}$ denote the ground-truth molecule sequence. The objective minimizes the negative log-likelihood of the molecule tokens given the prompt:

\begin{equation}
\mathcal{L}_{\text{CE}} = - \sum_{j=1}^{l} \log p_\theta(t_j^* \mid c, t_1^*, \ldots, t_{j-1}^*).
\end{equation}

Here, $p_\theta$ denotes the token-level predictive distribution of the model parameterized by $\theta$, and $t_j^*$ is the $j$-th token in the ground-truth molecule. The model is trained to generate molecules autoregressively by maximizing the likelihood of each token conditioned on the prompt $c$ and its preceding tokens.

To support efficient training and parameter reuse, we apply Low-Rank Adaptation (LoRA)~\cite{hu2021lora} across both supervised fine-tuning and reinforcement learning stages. LoRA injects trainable low-rank matrices into attention and feed-forward layers while keeping the original backbone weights frozen. This enables memory-efficient training and simplifies deployment in distributed settings. All parameter updates are confined to LoRA modules, ensuring architectural consistency across learning stages and enabling task-specific specialization with minimal overhead.

\subsection{Reinforcement learning with verifiable rewards}

Reinforcement learning has been widely adopted to enhance the reasoning capabilities of large language models (LLMs) following the supervised finetuning stage~\cite{ouyang2022training}. In this work, we extend Reinforcement Learning with Verifiable Rewards (RLVR) to the domain of molecular generation, where the correctness and utility of generated molecules can be explicitly verified via rule-based functions provided by cheminformatics toolkits such as RDKit~\cite{landrum2013rdkit}.

RLVR offers a simplified alternative to prior methods for bootstrapping LLM reasoning~\cite{zelikman2024quiet,hoffman2023training} and to RL frameworks that rely on execution feedback~\cite{gehring2024rlef}. Rather than using a learned reward model, RLVR leverages deterministic and binary verification signals—such as matching a correct answer or satisfying a structural or property-based constraint—as direct supervision. This makes RLVR particularly well-suited for molecular design tasks, where many objectives admit verifiable formulations. For example, molecule validity, property thresholds (e.g., QED $\geq$ 0.9), and similarity constraints (e.g., Tanimoto similarity $\geq$ 0.6) can all be checked deterministically. 

Building on the RLHF framework~\cite{ouyang2022training}, RLVR simplifies the reward mechanism by replacing the reward model with a verifiable reward function $v(c, m)$, which returns a scalar value if the generated molecule $m$ satisfies task-specific criteria given prompt $c$, and $0$ otherwise. The training objective is thus defined as:

\begin{equation}
\begin{split}
\max_{\pi_\theta} \mathbb{E}_{m \sim \pi_\theta(c)} \left[ R_{\text{RLVR}}(c, m) \right] \qquad\quad\quad\quad\quad\quad\quad\quad\quad\\
= v(c, m) - \beta D_{\text{KL}} \left[ \pi_\theta(m|c) \parallel \pi_{\text{ref}}(m|c) \right].
\end{split}
\end{equation}

\begin{figure*}[t]
\vspace{-8pt}
\centering
\begin{equation}
\label{eq:grpo_obj}
\begin{aligned}
\mathcal{J}_{\text{GRPO}}(\theta) = \mathbb{E}_{c \sim P(c), \{m_i\}_{i=1}^G \sim \pi_{\theta_{\text{old}}}} \left[ 
\frac{1}{G} \sum_{i=1}^{G} \frac{1}{|m_i|} \sum_{t=1}^{|m_i|} \mathcal{L}_{\text{token}}^{(i,t)} 
- \beta D_{\text{KL}}[\pi_\theta \| \pi_{\text{ref}}] \right].
\end{aligned}
\end{equation}

\vspace{-4pt}

\begin{equation}
\label{eq:grpo_token}
\mathcal{L}_{\text{token}}^{(i,t)} = 
\min \left(
\frac{\pi_\theta(m_{i,t} | c, m_{i,<t})}{\pi_{\theta_{\text{old}}}(m_{i,t} | c, m_{i,<t})} \hat{A}_{i,t},
\text{clip} \left(
\frac{\pi_\theta(m_{i,t} | c, m_{i,<t})}{\pi_{\theta_{\text{old}}}(m_{i,t} | c, m_{i,<t})}, 1 - \epsilon, 1 + \epsilon
\right) \hat{A}_{i,t}
\right).
\end{equation}
\label{fig:grpo_objective}
\vspace{-8pt}
\end{figure*}

\subsection{Group relative policy optimization}

To optimize the RLVR objective efficiently in the high-dimensional molecular generation space, we adopt the Group Relative Policy Optimization (GRPO) algorithm~\cite{shao2024deepseekmath}. GRPO is a lightweight and scalable variant of Proximal Policy Optimization (PPO)~\cite{schulman2017proximal} that removes the need for a separate value (critic) model. Instead, it estimates the learning baseline from the average reward of a group of sampled outputs, which aligns well with the binary reward signals commonly used in molecular verification. 
As illustrated in Stage 2 of Fig.~\ref{fig:framework}, for each input prompt corresponding to a specific task, GRPO samples a group of candidate molecules from the old policy, evaluates them using a verifiable reward function, and computes the relative advantages by subtracting the group's mean reward as a baseline for policy optimization.

As shown in Eq.~\ref{eq:grpo_obj}, the GRPO objective optimizes the policy by propagating a molecule-level scalar reward to token-level updates. This reward is averaged across a group of candidate molecules sampled under the same design condition to enable relative ranking. The inner token-level objective (Eq.~\ref{eq:grpo_token}) encourages the updated policy to increase the likelihood of favorable molecule completions, while clipping the probability ratio to stabilize training.
Here, $\hat{A}_{i,t}$ denotes the relative advantage for token $t$ in the $i$-th sampled molecule $m_i$, computed through outcome-level group-wise reward normalization. Specifically, for each molecular design condition $c$, a group of $G$ candidate molecules $\{m_1, \ldots, m_G\}$ is sampled from the old policy $\pi_{\theta_{\text{old}}}$, and their rewards $\{r_1, \ldots, r_G\}$ are obtained from a verifiable reward function $v(c, m_i)$. The normalized advantage is then calculated as:

\begin{equation}
\hat{A}_{i,t} = \tilde{r}_i = \frac{r_i - \text{mean}(r)}{\text{std}(r)}.
\end{equation}

This normalized scalar reward $\tilde{r}_i$ is assigned as the advantage $\hat{A}_{i,t}$ for all tokens in the sequence of molecule $m_i$. This design encourages comparisons among molecules generated under the same condition $c$, enabling stable optimization even with binary or sparse reward signals common in molecular design. This outcome-based advantage formulation avoids the need to train a separate value function, while preserving relative preference between candidate molecules.
The clipping mechanism in Eq.~\ref{eq:grpo_token} follows the PPO-style surrogate loss to prevent excessively large updates by restricting the policy ratio within the interval $[1 - \epsilon, 1 + \epsilon]$. The group size $G$ controls how many candidate molecules are sampled per prompt to compute the intra-group statistics, serving as a trade-off between computational efficiency and reward variance reduction.

To further regularize the policy, we apply a KL divergence penalty between the current policy and a frozen reference policy, which is initialized from the supervised finetuned model. Rather than incorporating the KL term into the reward function, we add it directly to the loss~\cite{shao2024deepseekmath}. This helps decouple the reward shaping from exploration control. The KL divergence is estimated using the unbiased estimator defined in Eq.~\ref{eq:kl_estimator}, and the coefficient $\beta$ controls its strength.

\begin{equation}
\label{eq:kl_estimator}
\begin{split}
D_{\text{KL}} \left[ \pi_\theta \| \pi_{\text{ref}} \right] = 
\frac{\pi_{\text{ref}}(m_{i,t} \mid c, m_{i,<t})}{\pi_\theta(m_{i,t} \mid c, m_{i,<t})} \quad\quad\quad\quad\quad\quad\\ 
- \log \frac{\pi_{\text{ref}}(m_{i,t} \mid c, m_{i,<t})}{\pi_\theta(m_{i,t} \mid c, m_{i,<t})} - 1,\quad
\end{split}
\end{equation}

which is guaranteed to be positive. This formulation simplifies optimization while reducing variance and computational overhead, making GRPO well-suited for molecular generation tasks, where reward signals are typically sparse and verifiable.

\subsection{Task-specific reward design}
Having introduced the GRPO optimization framework for reinforcement learning with verifiable rewards, we now detail the design of reward functions for various molecular generation objectives. Since molecular tasks can vary significantly in their property goals and constraint types, we employ task-specific verifiers to define meaningful and differentiable signals.

As a first step in reward computation, all completions are validated using RDKit. Invalid molecules are assigned a minimal reward as a penalty, which discourages the model from generating syntactically incorrect or chemically infeasible structures, and prevents RLVR training from collapsing due to the accumulation of invalid outputs.

\textbf{logP targeting:} In this task, the goal is to generate molecules whose physicochemical property---specifically logP---lies within a specified target range. To provide meaningful gradients and avoid sparse feedback, we design a reward function that assigns high scores to molecules within the target range, and applies a smooth decay for near-boundary cases.

Given a predicted logP value $\hat{p}$ for a generated molecule $m$, the reward function is defined as:

\begin{equation}
v_{\text{logP}}(c, m) =
\begin{cases}
1, & \text{if } p_{\text{min}} \leq \hat{p} \leq p_{\text{max}}, \\
\psi(\hat{p}), & \text{if } |\hat{p} - p_{\text{max}}| < \delta, \\
0, & \text{otherwise},
\end{cases}
\label{eq:logp_reward}
\end{equation}

where $p_{\text{min}}$ and $p_{\text{max}}$ specify the target logP interval, and $\delta$ controls the width of the smooth transition region. The smoothing function $\psi(\hat{p})$ is defined as:

\begin{equation}
\psi(\hat{p}) = \exp\left( - \frac{\min\left(|\hat{p} - p_{\text{min}}|, |\hat{p} - p_{\text{max}}|\right)}{\delta} \right).
\label{eq:logp_smooth}
\end{equation}

Molecules with $\hat{p}$ values slightly outside the desired range receive gradually decayed rewards, while those far outside (more than $\delta$ away) receive zero reward. This formulation promotes exploration while maintaining a smooth reward landscape, facilitating more stable and sample-efficient learning.

\textbf{QED and penalized logP maximization:} In this task, the goal is to generate molecules that maximize specific properties: QED and P-logP.
QED scores lie in the range $(0, 1)$ and exhibit a natural upper bound around $0.948$~\cite{zhou2019optimization}, due to the way they are computed from empirical distributions. Drug-like molecules with QED $\geq 0.9$ are extremely rare~\cite{bickerton2012quantifying}. To emphasize high QED regions while maintaining gradient signal, we square the score to obtain the reward:
\begin{equation}
v_{\text{QED}}(c, m) = \text{RDKit.QED}(m)^2.
\label{eq:qed_reward}
\end{equation}
 
The penalized logP metric accounts for the octanol–water partition coefficient, synthetic accessibility, and structural features. However, as noted in prior work~\cite{zhou2019optimization}, P-logP can be trivially increased by generating long alkyl chains. To mitigate this issue, we apply a hybrid logarithmic transformation to compress the scale and reduce the reward bias towards large molecules. The reward is defined as:

\begin{equation}
v_{\text{P-logP}}(c, m) =
\begin{cases}
\log(1 + s), & \text{if } s \geq 0, \\
- \log(1 + |s|), & \text{if } s < 0,
\end{cases}
\label{eq:plogp_reward}
\end{equation}
where $s = \text{RDKit.P-logP}(m)$. This hybrid design ensures that both positive and negative scores are smoothly compressed, discouraging degenerate molecule structures while preserving relative differences across the reward spectrum.

\textbf{Similarity-constrained target property maximization:}  
In this task, the objective is to generate molecules that simultaneously maintain structural similarity to a reference compound and exhibit improved molecular properties, such as penalized logP or QED. This setting closely aligns with real-world drug discovery scenarios like lead optimization, where newly designed candidates are expected to retain key structural motifs from known actives while enhancing specific physicochemical attributes.

Let $s(m)$ denote the Tanimoto similarity between the reference molecule $m_\text{ref}$ and the generated molecule $m$:
\begin{equation}
s(m) = \text{Tanimoto similarity}(m_\text{ref}, m).
\end{equation}

We define the similarity reward as:
\begin{equation}
v_{\text{sim}}(c, m) = s(m)^2.
\end{equation}

The final reward function is constructed as a weighted sum of the similarity component and the target property score (either penalized logP or QED), denoted generically as $v_{\text{prop}}(c, m)$:
\begin{equation}
v_{\text{joint}}(c, m) = \lambda \cdot v_{\text{sim}}(c, m) + (1 - \lambda) \cdot v_{\text{prop}}(c, m).
\label{eq:joint_reward}
\end{equation}

Here, $v_{\text{prop}}(c, m)$ denotes either $v_{\text{P-logP}}(c, m)$ or $v_{\text{QED}}(c, m)$ depending on the task, as defined in Eq.~\ref{eq:plogp_reward} and Eq.~\ref{eq:qed_reward}, respectively.
The hyperparameter $\lambda \in [0,1]$ governs the trade-off between preserving molecular similarity and enhancing target properties. This reward formulation encourages the generation of molecules that remain within the local chemical space of the reference compound while progressively improving target characteristics.
Notably, the modular structure of this reward design highlights the scalability of the RLVR framework with GRPO, allowing seamless extension to multi-objective optimization tasks involving diverse molecular properties.

\section{Experiments}
In this section, we detail the experimental setup used to evaluate our proposed framework, LLMol, and present comprehensive results across multiple molecular generation tasks introduced in Section~\ref{sec: task-description}. Our primary goal is to address the following questions:

\begin{itemize}
\item \textbf{Q1)} Can LLMol learn the syntactic and semantic structure of molecular data and generate valid, novel molecules through supervised finetuning?
\item \textbf{Q2)} Does LLMol achieve higher success rates in generating molecules that fall within the desired logP range?
\item \textbf{Q3)} Can LLMol generate molecules with improved QED and penalized logP scores compared to baseline models?
\item \textbf{Q4)} In the similarity-constrained setting, can LLMol effectively improve molecular properties while preserving structural similarity to known compounds?
\end{itemize}

\subsection{Experimental Setup}

We conduct experiments on the ZINC dataset following the LIMO protocol~\cite{eckmann2022limo}. In the first stage, supervised fine-tuning, we randomly sample over 200,000 molecules from the publicly available ZINC250K dataset~\cite{wu2018moleculenet}. The dataset is randomly split into training and test sets using a $9{:}1$ ratio.
For model initialization, we adopt Qwen-3~\cite{qwen3technicalreport}, a 4-billion-parameter language model, as the base model for supervised fine-tuning. Training is implemented using the Transformers library~\cite{wolf2020transformers} and PyTorch~\cite{paszke2019pytorch}. 
To facilitate efficient training on limited GPU resources, we apply LoRA~\cite{hu2021lora} to reduce the number of trainable parameters.

\subsection{Baselines}

We consider a diverse set of baseline models spanning four major categories: VAE-based methods (e.g., JT-VAE~\cite{jin2018junction}, LIMO~\cite{eckmann2022limo}), reinforcement learning-based approaches (e.g., GCPN~\cite{you2018graph}, MOLGQN~\cite{zhou2019optimization}, GraphDF~\cite{luo2021graphdf}), autoregressive and flow-based models (e.g., GraphAF~\cite{shi2020graphaf}), and Transformer-based molecular language models (e.g., Chemformer~\cite{irwin2022chemformer}, RT~\cite{born2023regression}, MOLGEN~\cite{fang2024domain}). We also include retrieval-augmented generation (RetMol~\cite{wang2022retrieval}) and MCMC-based frameworks (MARS~\cite{xie2021mars}).
Details of each baseline are provided in Appendix~\ref{appendix:baselines}.

\begin{table}[!t]
\renewcommand\arraystretch{1.2}  
\centering
\small
\caption{The performance of supervised learning on ZINC. Validity and novelty are reported as percentages. Baseline results taken from~\cite{fang2024domain}.}
\label{tab:sft}
\begin{tabular}{lccc}
\toprule
\textbf{Method} & \textbf{Validity~$\uparrow$} & \textbf{Diversity~$\uparrow$} & \textbf{Novelty~$\uparrow$} \\
\midrule
AAE          & 93.68\% & 0.8557 & 79.31\% \\
LatentGAN    & 89.66\% & \cellcolor{lightblue2}\textbf{0.8565} & 94.98\% \\
CharRNN      & 97.48\% & 0.8562 & 84.19\% \\
VAE          & 97.67\% & 0.8558 & 69.49\% \\
JT-VAE       & \cellcolor{lightblue2}99.65\% & 0.8551 & 91.43\% \\
LIMO         & \cellcolor{lightblue1}\textbf{100.00\%} & 0.8544 & 89.56\% \\
Chemformer   & 98.43\% & 0.8553 & \cellcolor{lightblue2}95.81\% \\
MOLGEN       & \cellcolor{lightblue1}\textbf{100.00\%} & \cellcolor{lightblue1}\textbf{0.8567} & \cellcolor{lightblue1}\textbf{100.00\%} \\
\midrule
LLMol$^\dagger$   & \cellcolor{lightblue1}\textbf{100.00\%} & 0.8473 & \cellcolor{lightblue1}\textbf{100.00\%} \\
\bottomrule
\end{tabular}
\vspace{-8pt}
\end{table}

\subsection{Evaluation}

To evaluate the quality of generated molecules, we adopted the metrics following baseline MOLGEN~\cite{fang2024domain}. 
\textbf{Validity} measures the proportion of generated molecules that conform to chemical valence rules, ensuring structural correctness.
\textbf{Diversity} evaluates the chemical diversity within the generated set by computing the average pairwise Tanimoto distance, reflecting the model's ability to produce varied molecular structures.
\textbf{Novelty} quantifies the percentage of generated molecules that are not present in the training set, indicating the model's capacity to explore novel regions of chemical space.

\begin{table}[b]
\vspace{-8pt}
\centering
\small
\caption{Property targeting to $-2.5 < logP < -2.0$. Success (\%): percent of generated molecules within the target range. Diversity: One minus the average pairwise Tanimoto similarity between Morgan fingerprints. Baseline results taken from~\cite{eckmann2022limo}}
\label{tab:logp-targeting}
\begin{tabular}{lcc}
\toprule
\textbf{Method} & \textbf{Success (\%)} & \textbf{Diversity} \\
\midrule
ZINC250K & 0.44 & 0.919 \\
\midrule
JT-VAE & 11.3 & 0.846 \\
GCPN & \cellcolor{lightblue1}\textbf{85.5} & 0.392 \\
MolDQN & 9.66 & 0.854 \\
GraphDF & 0 & - \\
MOLGEN & 0 & - \\
LIMO & 10.4 & \cellcolor{lightblue1}\textbf{0.914} \\

\midrule
LLMol \dag      & 0.46 & 0.887 \\
LLMol       & \cellcolor{lightblue2}46.91 & \cellcolor{lightblue2}0.893 \\
\bottomrule
\end{tabular}
\end{table}

\subsection{Results and discussion}

\subsubsection{Q1: Performance on molecular distribution learning}

A foundational requirement for molecular generation models is the ability to capture the structural and semantic distribution of molecules and generate diverse, valid, and novel candidates. This capacity is critical for constructing virtual libraries in computer-aided drug discovery~\cite{van2019virtual}, where the quality and variability of generated molecules directly affect downstream screening and optimization stages.
To evaluate the generative capacity of our supervised model, LLMol$^\dagger$, we compare it against established baselines on the ZINC dataset. The results are summarized in Table~\ref{tab:sft}. 

LLMol$^\dagger$ achieves 100\% validity—matching the strongest baselines. This demonstrates that a general-purpose language model, when fine-tuned on simplified SELFIES representations, can effectively learn the syntactic and valence constraints of molecular structures without requiring explicit graph-based modules or large-scale molecular pretraining.
In terms of novelty, LLMol$^\dagger$ matches MOLGEN with a perfect score (100\%), highlighting its ability to explore novel regions of the chemical space beyond the training distribution. This suggests that fine-tuned LLMs can generalize molecular syntax in a way that supports both correctness and creative generation. While diversity is slightly lower, LLMol$^\dagger$ still achieves competitive performance. This minor gap can be attributed to the balance between validity enforcement and chemical exploration during autoregressive decoding.
Overall, LLMol$^\dagger$ demonstrates a strong ability to model molecular distributions with high validity and novelty, using only prompt-based supervised tuning—without the need for domain-specific inductive biases or pretraining on large-scale molecular corpora.

\begin{table*}[!t]
\small
\caption{Comparison of QED and Penalized logP maximization methods. "LL" (length limit) denotes whether a model has a limited output length (about the maximum molecule size of ZINC250k), as P-logP score can increase linearly with molecule length. Baseline results taken from~\cite{eckmann2022limo, fang2024domain }\label{tab:plogp-qed}} 
\centering
\setlength\tabcolsep{9.5pt}
\begin{tabular}{l c ccc ccc c}
\toprule
\textbf{Method} & \textbf{LL} & \multicolumn{3}{c}{\textbf{Penalized logP}} & \multicolumn{3}{c}{\textbf{QED}} & \textbf{Time} \\
\cmidrule(lr){3-5} \cmidrule(lr){6-8}
& & 1\text{st} & 2\text{nd} & 3\text{rd} & 1\text{st} & 2\text{nd} & 3\text{rd} & (HRS)\\
\midrule
ZINC250K      &  & 4.52  & 4.30  & 4.23  & 0.948 & 0.948 & 0.948 &  \\
\midrule
JT-VAE           & \ding{55} & 5.30  & 4.93  & 4.49  & 0.925 & 0.911 & 0.910 & 24 \\
MARS             & \ding{55} & \cellcolor{lightblue1}\textbf{45.0} & \cellcolor{lightblue1}\textbf{44.3} & \cellcolor{lightblue1}\textbf{43.8} & \cellcolor{lightblue1}\textbf{0.948} & \cellcolor{lightblue1}\textbf{0.948} & \cellcolor{lightblue1}\textbf{0.948} & 12 \\
GraphDF          & \ding{55} & 13.7  & 13.2  & 13.2  & \cellcolor{lightblue1}\textbf{0.948} & \cellcolor{lightblue1}\textbf{0.948} & \cellcolor{lightblue1}\textbf{0.948} & 8 \\
GraphAF          & \ding{55} & 12.23 & 11.29 & 11.05 & \cellcolor{lightblue1}\textbf{0.948} & \cellcolor{lightblue1}\textbf{0.948} & \cellcolor{lightblue2}0.947 & 8\\
\midrule
GCPN             & \ding{51} & 7.98  & 7.85  & 7.80  & \cellcolor{lightblue1}\textbf{0.948} & \cellcolor{lightblue2}0.947 & 0.946 & 8 \\
MolDQN           & \ding{51} & 11.8  & 11.8  & 11.8  & \cellcolor{lightblue1}\textbf{0.948} & 0.943 & 0.943 & 24 \\
MOLGEN            & \ding{51} & 30.52 & 28.73 & 28.65 & \cellcolor{lightblue1}\textbf{0.948} & \cellcolor{lightblue2}0.947 & 0.946 & 16\\
LIMO             & \ding{51} & 10.5  & 9.69  & 9.60  &\cellcolor{lightblue2} 0.947 & 0.946 & 0.945 & \cellcolor{lightblue1}\textbf{1} \\
\midrule
LLMol \dag      & \ding{51} & 4.86  & 4.59  & 4.24  & \cellcolor{lightblue1}\textbf{0.948} & 0.944 & 0.942 & \cellcolor{lightblue2}\textbf{2} \\
LLMol           & \ding{51} & \cellcolor{lightblue2}{31.12} & \cellcolor{lightblue2}{30.05} & \cellcolor{lightblue2}29.96    & \cellcolor{lightblue1}\textbf{0.948} & \cellcolor{lightblue1}\textbf{0.948} & \cellcolor{lightblue1}\textbf{0.948} & \cellcolor{lightblue1}\textbf{1} \\

\bottomrule
\end{tabular}
\end{table*}

\subsubsection{Q2: Performance on logP targeting}

Table~\ref{tab:logp-targeting} reports the performance of LLMol on the task of generating molecules whose logP values fall within the narrow target interval of $-2.5 < \text{logP} < -2.0$. This task poses a significant challenge due to the sparsity of valid molecules in this range and the need for precise property-controlled generation. Such constraints often reflect early-stage drug-likeness filters related to solubility and permeability.
We evaluate two variants of our model: LLMol$^\dagger$, trained only via supervised finetuning, and the full LLMol, which additionally incorporates reinforcement learning with verifiable reward optimization. The SFT-only model achieves a success rate of 0.46\%, which closely mirrors the empirical distribution in ZINC250K (0.44\%). This suggests that the supervised model faithfully reproduces the training data distribution, without introducing directional bias in logP values.

In contrast, the RL-augmented model achieves a significantly higher success rate of 46.91\%, surpassing all prior baselines by a substantial margin. This demonstrates that incorporating verifiable reward signals via RLVR enables the model to generate molecules that satisfy narrow physicochemical constraints with high precision.
Beyond property satisfaction, LLMol also maintains high structural diversity: it achieves a diversity score of 0.893, approaching that of LIMO (0.914). This indicates that reinforcement learning does not compromise the model's ability to explore a chemically rich and varied molecular space.

\subsubsection{Q3: Performance on QED and penalized logP maximization}

Targeted molecular design focuses on generating novel molecules with optimized physicochemical properties. Following the evaluation protocol~\cite{shi2020graphaf,eckmann2022limo}, we report the top-3 penalized logP and QED scores in Table~\ref{tab:plogp-qed}. Since penalized logP increases with molecule length~\cite{zhou2019optimization}, we group baselines by whether they impose length limits (LL) to avoid inflation of reward scores.

\begin{figure*}[!t]
  \centering
  \includegraphics[width=\textwidth]{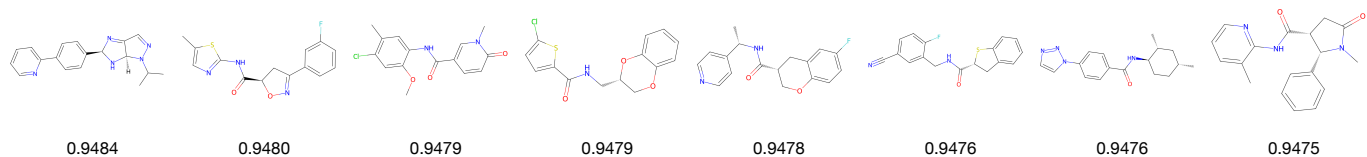}
  \caption{Six sampled molecules with the highest QED scores discovered by LLMol.}
  \label{fig:highest_QED}
\end{figure*}

LLMol$^\dagger$ achieves penalized logP scores comparable to those observed in the training dataset. In contrast, after applying reinforcement learning with verifiable rewards, LLMol achieves substantially higher scores, outperforming all prior methods. In terms of drug-likeness, LLMol demonstrates a strong ability to generate molecules with high QED, with 1.20\% of generated molecules surpassing a QED score of $0.9$, highlighting its effectiveness in optimizing specific molecular properties. 
These results indicate that LLMol is highly capable of targeted property optimization, achieving state-of-the-art results in a short generation time, and further solidifying its potential in efficient molecular design. Fig.~\ref{fig:highest_QED} shows six sampled molecules with the highest score discovered by our model.

\begin{table}[!t]
\centering
\small
\caption{Mean (and standard deviation) penalized logP improvement of generated molecules compared to inputs with different similarity constraints. Baseline results taken from~\cite{fang2024domain }.\label{tab:plogp-improvement}}
\begin{tabular}{@{}lcc@{}}
\toprule
\textbf{Method} & \multicolumn{2}{c}{\textbf{Improvement}} \\
\cmidrule(lr){2-3}
& \textbf{$\delta = 0.6$} & \textbf{$\delta = 0.4$} \\
\midrule
JT-VAE & 0.28 \, {\scriptsize(0.79)} & 1.03 \, {\scriptsize(1.39)} \\
GCPN & 0.79 \, {\scriptsize(0.63)} & 2.49 \, {\scriptsize(1.30)} \\
MolDQN & 1.86 \, {\scriptsize(1.21)} & 3.37 \, {\scriptsize(1.62)} \\
VSeq2Seq & 2.33 \, {\scriptsize(1.17)} & 3.37 \, {\scriptsize(1.75)} \\
VJTNN & 2.33 \, {\scriptsize(1.24)} & 3.55 \, {\scriptsize(1.67)} \\
GA & 3.44 \, {\scriptsize(1.09)} & 5.93 \, {\scriptsize(1.41)} \\
GraphAF & 4.98 \, {\scriptsize(6.49)} & 8.21 \, {\scriptsize(6.51)} \\
GraphDF & 4.51 \, {\scriptsize(5.80)} & 9.19 \, {\scriptsize(6.43)} \\
LIMO & 1.80 \, {\scriptsize(2.00)} & 3.60 \, {\scriptsize(2.30)} \\
Chemformer & 2.48 \, {\scriptsize(0.89)} & 3.56 \, {\scriptsize(1.32)} \\
RetMol & 3.78 \, {\scriptsize(3.29)} & 11.55 \, {\scriptsize(11.27)} \\
RT & 2.21 \, {\scriptsize(1.30)} & 3.16 \, {\scriptsize(1.50)} \\
MOLGEN & \cellcolor{lightblue1}12.08 \, {\scriptsize(0.82)} & \cellcolor{lightblue2}12.35 \, {\scriptsize(1.21)} \\
\midrule
\textbf{LLMol} & \cellcolor{lightblue2}10.60 \, {\scriptsize(13.00)} & \cellcolor{lightblue1}17.50 \, {\scriptsize(13.82)} \\

\bottomrule
\end{tabular}
\vspace{-8pt}
\end{table}

\begin{figure*}[t]
  \centering
  \includegraphics[width=\textwidth]{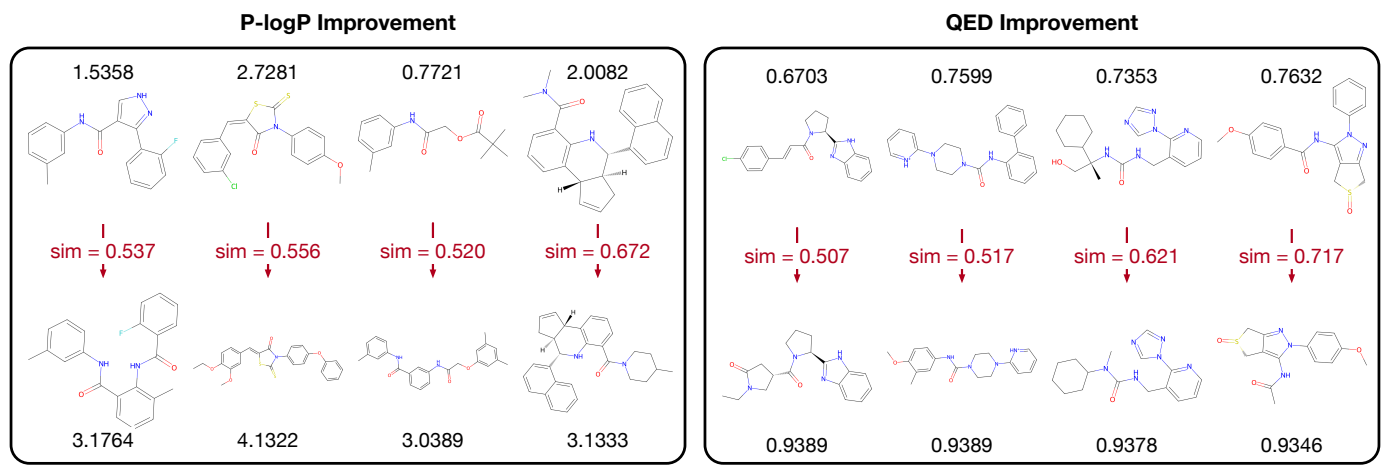}
  \caption{An illustration of constrained optimization results based on P-logP and QED scores. The top row shows reference molecules from the test dataset, while the bottom row displays the molecules generated by LLMol. The value \textit{sim} denotes the Tanimoto similarity between the reference and the generated molecule.}
  \label{fig:sim-constraint-examples}
  \vspace{-8pt}
\end{figure*}

\subsubsection{Q4: Similarity-constrained target property maximization}

This optimization task aims to enhance specific molecular properties while preserving structural similarity to a reference molecule. Following MOLGEN~\cite{fang2024domain}, we optimize 800 molecules from the ZINC250K dataset with the lowest penalized logP scores, under different similarity constraints (denoted as $\delta$). The similarity between the original and optimized molecules is assessed using the Tanimoto coefficient computed over Morgan fingerprints~\cite{rogers2010extended}.

As shown in Table~\ref{tab:plogp-improvement}, LLMol outperforms baseline models under both similarity thresholds, demonstrating its capacity to generate molecules with improved properties while staying close in chemical space to their reference compounds. These results validate the effectiveness of LLMol in similarity-constrained optimization, a setting critical to applications such as lead optimization in drug discovery.

In addition, Fig.~\ref{fig:sim-constraint-examples} illustrates qualitative examples of similarity-constrained optimization for both penalized logP and QED. The top row displays reference molecules, while the bottom row presents generated molecules with improved scores. The annotated similarity values confirm that the generated molecules remain structurally similar to their respective references while achieving significant property enhancements.

\section*{Conclusion}

In this work, we present a novel molecular design framework based on Reinforcement Learning with Verifiable Rewards and Group Relative Policy Optimization. Our approach combines supervised fine-tuning of large language models with reinforcement learning, utilizing RDKit-based rewards for efficient and objective optimization of molecular properties.
Experimental results demonstrate that our method outperforms state-of-the-art baselines in generating molecules with target properties while maintaining structural similarity constraints. Additionally, our framework achieves high efficiency, significantly reducing computational time compared to existing methods.
Overall, RLVR with GRPO provides a scalable and effective solution for targeted molecular optimization. Future work will focus on enhancing model scalability and exploring more complex molecular design tasks.

\bibliographystyle{IEEEtran}
\bibliography{main}
\newpage
\vfill

\section{Appendix}

\subsection{Baseline Descriptions}
\label{appendix:baselines}
\textbf{JT-VAE}~\cite{jin2018junction} is a VAE-based generative model that first generates a scaffold junction tree and then assembles its nodes into a complete molecular graph.

\textbf{GCPN}~\cite{you2018graph} is an RL-based molecular graph generation method that optimizes molecules by maximizing a composite reward function incorporating both adversarial loss and property-specific objectives.

\textbf{MOLGQN}~\cite{zhou2019optimization} is an RL-based method that integrates double Q-learning with chemical domain knowledge to guide molecular generation more effectively.

\textbf{MARS}~\cite{xie2021mars} is a Markov Chain Monte Carlo sampling framework that employs an adaptive fragment-editing proposal distribution, enhanced with Graph Neural Networks to propose chemically valid modifications.

\textbf{GraphAF}~\cite{shi2020graphaf} is an autoregressive flow-based model that constructs molecular graphs by sequentially adding nodes and edges in a learnable, probabilistic manner.

\textbf{GraphDF}~\cite{luo2021graphdf} is a normalizing flow model for molecule generation that operates in a discrete latent space and is further refined via reinforcement learning.

\textbf{Chemformer}~\cite{irwin2022chemformer} is a pre-trained Transformer-based molecular language model that operates on SMILES representations for both generative and discriminative tasks.

\textbf{LIMO}~\cite{eckmann2022limo} is a variational autoencoder (VAE)-based framework that explores a latent space tailored for molecular design, enabling controlled molecule optimization and generation.

\textbf{RetMol}~\cite{wang2022retrieval} is a retrieval-augmented generation framework built on Chemformer, which leverages a task-specific molecular database to guide generation toward satisfying predefined design constraints.

\textbf{RT}~\cite{born2023regression} is a Transformer-based model pre-trained on SELFIES. It generates molecules by taking expected molecular property values as input along with a given molecular scaffold, ensuring that the generated molecules incorporate this scaffold. 

\textbf{MOLGEN}~\cite{fang2024domain}, a pre-trained molecular language model on SELFIES, employs prefix tuning to enhance generation diversity and utilizes contrastive learning to ensure alignment with predefined chemical preferences.

\subsection{Prompt Design}
\label{appendix:prompt}

We design task-specific prompts for both supervised fine-tuning (SFT) and reinforcement learning with verifiable rewards (RLVR). Prompts are constructed in a structured, instruction-driven format to facilitate reliable molecule generation under property or constraint guidance.

Each training instance consists of: (i) a system message defining the assistant's role, (ii) an illustrative example with a molecule and its associated property value, and (iii) a generation instruction. For SFT, the prompt is followed by a ground-truth response. Examples are randomly sampled from the training set to enhance diversity. For RLVR, we adopt a similar format but revise the instruction to reflect optimization intent (e.g., "as high as possible"). Optionally, a high-quality reference example can be included to stabilize early-stage policy learning.

Below we show one prompt used for logP-targeted SFT, and one for QED maximization in the RLVR setting:

\begin{tcolorbox}[title=Prompt for logP Targeting (SFT), colback=gray!5, colframe=black!40]
\begin{lstlisting}
[{
    "role": "system",
    "content": "You specialize in molecular design and property-constrained molecule generation."
  },{
    "role": "user",
    "content": "Example:\nProperty: logP\nValue: <value>1.90</value>\n<SELFIES> C C Branch1 C C NH2+1 C C =C O C =C C =C C =C Ring1 =Branch2 Ring1 =Branch1 </SELFIES>"
  },{
    "role": "user",
    "content": "Now generate a molecule with:\nProperty: logP\nValue: <value>1.16</value>"
  },{
    "role": "assistant",
    "content": "<SELFIES> C C C@@H1 Branch1 C C C O C@@H1 C C C NH2+1 C@@H1 Ring1 =Branch1 C </SELFIES>"
  }]
\end{lstlisting}
\end{tcolorbox}

\begin{tcolorbox}[title=Prompt for QED Maximization (RLVR), colback=gray!5, colframe=black!40]
\begin{lstlisting}
[{
    "role": "system",
    "content": "You specialize in molecular design and property-constrained molecule generation.\nYour goal is to generate molecules with as high QED as possible."
  },{
    "role": "user",
    "content": "Example:\nProperty: QED\nValue: <value>0.946</value>\n<SELFIES> C C N C =C Branch1 C Br C Branch1 =C C =Branch1 C =O N C =C C =C C =N Ring1 =Branch1 =N Ring1 #C </SELFIES>"
  },{
    "role": "user",
    "content": "Now generate a molecule with:\nProperty: QED\nValue: <value>as high as possible</value>"
  }]
\end{lstlisting}
\end{tcolorbox}

All prompts are serialized using the $apply\_chat\_template$ function of the Qwen3 tokenizer~\cite{qwen3technicalreport}, ensuring consistent formatting across both SFT and RLVR training regimes.

\subsection{Model Training}
\label{appendix:training}

This section provides detailed hyperparameter configurations for both the supervised fine-tuning (SFT) stage and the reinforcement learning with verifiable rewards (RLVR) stage.

\begin{table}[!t]
\centering
\small
\caption{Hyperparameter configurations for LLMol training.}
\label{tab:hyperparameters}
\renewcommand{\arraystretch}{1.2}
\begin{tabular}{@{}ll@{}}
\toprule
\textbf{Hyperparameter} & \textbf{Value} \\
\midrule
\multicolumn{2}{l}{\textit{Model Configuration}} \\
\midrule
Base model & Qwen-3 (4B) \\
LoRA rank & 16 \\
LoRA scaling factor & 32 \\
LoRA dropout & 0.05 \\
\midrule
\multicolumn{2}{l}{\textit{Supervised Fine-tuning}} \\
\midrule
Training epochs & 8 \\
Learning rate & $1 \times 10^{-6}$ \\
Per-device batch size & 2 \\
Gradient accumulation steps & 8 \\
Effective batch size & 16 \\
Gradient checkpointing & Enabled \\
\midrule
\multicolumn{2}{l}{\textit{Reinforcement Learning (GRPO)}} \\
\midrule
Training epochs & 8 \\
Learning rate & $1 \times 10^{-6}$ \\
Per-device batch size & 4 \\
Gradient accumulation steps & 4 \\
KL coefficient ($\beta$) & 0.2 \\
Group size ($G$) & 16 \\
Top-$k$ sampling & $k = 20$ \\
Nucleus sampling ($p$) & 0.9 \\
Temperature & 0.8 \\
Max prompt length & 512 tokens \\
Max completion length & 128 tokens \\
Mixed precision & bfloat16 \\
\bottomrule
\end{tabular}
\end{table}

All experiments are conducted using the HuggingFace Transformers library~\cite{wolf2020transformers} and PyTorch~\cite{paszke2019pytorch}. The supervised fine-tuned model is used to initialize both the policy model and the frozen reference model for GRPO optimization. Task-specific verifiable reward functions are implemented using RDKit~\cite{landrum2013rdkit}.

\begin{figure*}[t]
\centering
\includegraphics[width=\textwidth]{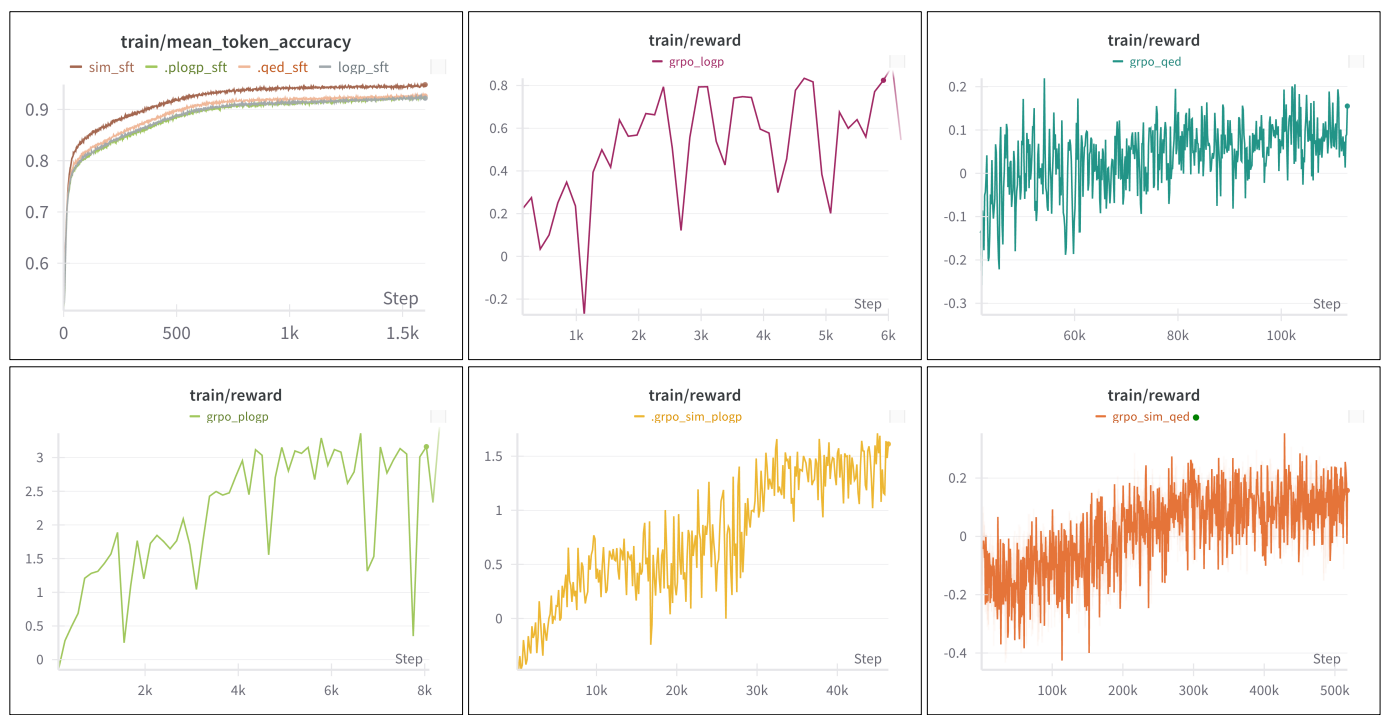}

\caption{
Training curves across different tasks. Each plot shows the metric trend over training steps. The top-left figure shows the supervised fine-tuning (SFT) process on three molecular properties and similarity, while the remaining plots illustrate reward progression during Reinforcement Learning with Verifiable Rewards (RLVR) with Group Relative Policy Optimization (GRPO).
}
\label{fig:allcarve}
\end{figure*}

\end{document}